\documentclass[11pt,a4paper]{article}
\usepackage{times,latexsym}
\usepackage{url}
\usepackage[T1]{fontenc}
\usepackage{enumitem}
\usepackage[acceptedWithA]{tacl2021v1}
\usepackage{xspace,mfirstuc,tabulary}
\usepackage{booktabs}
\usepackage{graphicx}
\usepackage{amsfonts} 
\usepackage{amsthm,amsmath,amssymb,amsfonts}
\usepackage[most]{tcolorbox}
\usepackage{multirow}
\usepackage{adjustbox}

\newcommand{\stitle}[1]{\vspace{1ex}\noindent{\bf #1}}

\newif\iftaclinstructions
\taclinstructionsfalse % AUTHORS: do NOT set this to true
\iftaclinstructions
\renewcommand{\confidential}{}
\renewcommand{\anonsubtext}{(No author info supplied here, for consistency with
TACL-submission anonymization requirements)}
\newcommand{\instr}
\fi

\iftaclpubformat % this "if" is set by the choice of options

\else

\fi

\title{%RAG’s Achilles’ Heel: When Similarity Search Misses the Big Picture Domain-level Compressed Corpora To The Rescue 
Beyond RAG: Task-Aware KV Cache Compression\\ for Comprehensive Knowledge Reasoning}

\author{
Giulio Corallo  \\
  SAP Labs, \\
  EURECOM  \\
  \texttt{\footnotesize giulio.corallo@sap.com}
  \And
  Orion Weller\\
  Johns Hopkins University\\
  \texttt{\footnotesize oweller@cs.jhu.edu}
  \And
  Fabio Petroni\\
  Samaya AI\\
  \texttt{\footnotesize fabio@samaya.ai}\\
  \And
Paolo Papotti  \\
EURECOM\\
  \texttt{ \footnotesize papotti@eurecom.fr}\\
}

\date{}
\begin{document}
\maketitle
\begin{abstract}
Incorporating external knowledge in large language models (LLMs) enhances their utility across diverse applications, but existing methods have trade-offs. Retrieval-Augmented Generation (RAG) fetches evidence via similarity search, but key information may fall outside top ranked results. Long-context models can process multiple documents but are computationally expensive and limited by context window size.
%Incorporating external information in large language models (LLMs) is crucial for enhancing their utility across diverse applications. Retrieval-Augmented Generation (RAG) leverages external corpora to retrieve relevant information, while long-context models allow for the direct processing. %, bypassing retrieval altogether. 
%However, both approaches have shortcomings, such as %the limitations due to
%incomplete evidence retrieval in the former and the high computational requirements for the latter. 

Inspired by students condensing study material for open-book exams, we propose task-aware key-value (KV) cache compression, which compresses external knowledge in a zero- or few-shot setup. This enables LLMs to reason efficiently over a compacted representation of all relevant information.

Experiments show our approach outperforms both RAG and task-agnostic compression methods. On LongBench v2, it improves accuracy by up to 7 absolute points over RAG with a 30× compression rate, while reducing inference latency from 0.43s to 0.16s. A synthetic dataset highlights that RAG performs well when sparse evidence suffices, whereas task-aware compression is superior for broad knowledge tasks.

\end{abstract}

\section{Introduction}

Incorporating external information into large language models (LLMs) significantly enhances their utility across various applications, enabling them to generate more informed and accurate outputs. Several methodologies have been devised to facilitate this integration, but each comes with limitations that restrict its effectiveness.

\begin{figure}[t]
    \centering
    \includegraphics[width=\columnwidth]{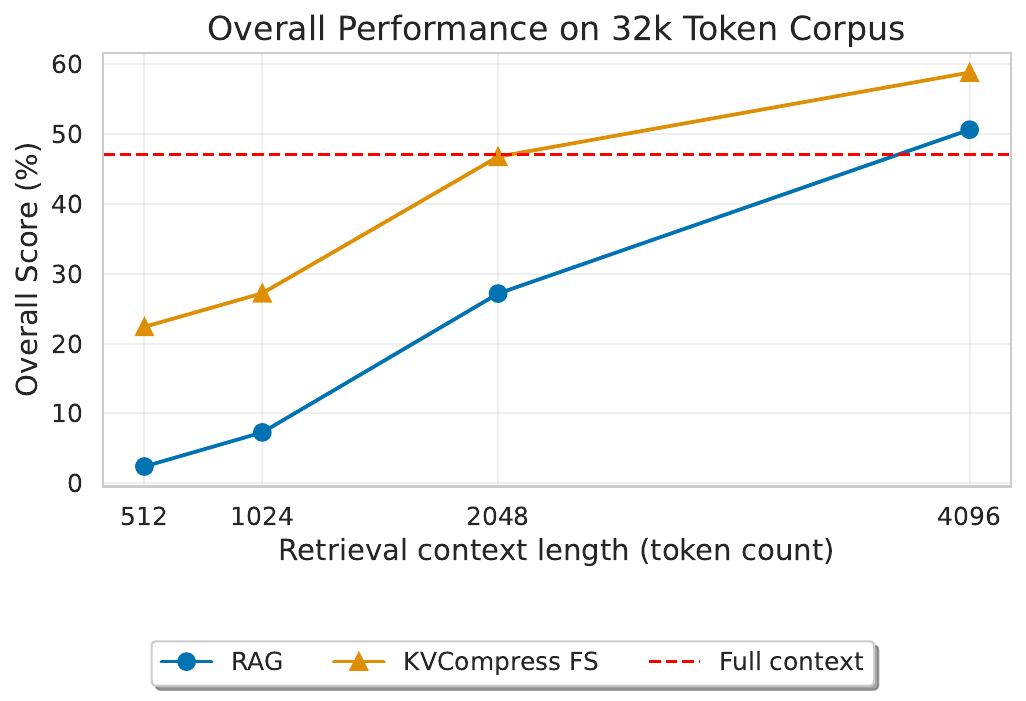}
    \caption{Overlap match of the words between ground truth and predictions of various KV cache compression methods compared to RAG on a synthetic corpus with 32k tokens. Our Few-Shot compression approach achieves results exceeding RAG when the context length is much smaller than the corpus size.}
    \label{fig:main_image}
\end{figure}

Retrieval-Augmented Generation (RAG) is a technique that enhances LLMs by leveraging external corpora to retrieve relevant chunks of information~\citep{lewis2020retrieval}. However, RAG is most effective in scenarios with narrow, focused queries that require a few pieces of evidence. When dealing with broader queries that demand synthesizing insights from multiple sources across the corpus, retrieval mechanisms may fall short~\citep{barnett2024seven}. This happens because retrieval typically relies on similarity-based search, which may fail to capture implicit relationships between distant pieces of evidence, making it challenging to surface all relevant context and often introducing noise or redundancy~\citep{yu2024knowledge}.
Researchers have begun to address these issues, e.g., via improved chunking and pruning strategies%or graphs
~\citep{edge2024local}, but handling broad queries with RAG remains challenging.
%Additionally, retrieval methods often impose a hard constraint on the number of retrieved documents, making it challenging to surface all relevant context and often introducing noise or redundancy.

\begin{figure*}[t]
    \centering
\includegraphics[width=\textwidth]{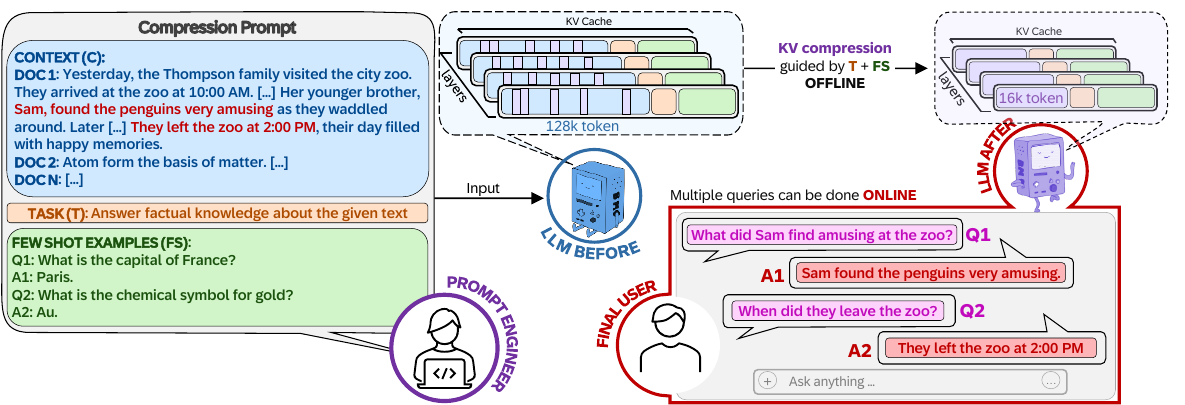}
    \caption{An illustration of our compression strategy that reduces the original context (C) from a KV cache  of 128k tokens to 16k. This process is guided by task instructions (T) and few-shot examples (FS), condensing the essential information needed for factual QA on the corpus documents. At inference time, the LLM can answer multiple questions as if it had access to the entire (uncompressed) corpus. %[stress that we can answer many questions after compression].
    }
    \label{fig:intro}
\end{figure*}

Recent advancements have extended LLMs’ ability to process longer contexts, pushing the boundaries of how much information they can handle simultaneously~\citep{team2024gemini, li2025minimax}. This progress opens up the possibility of processing entire corpora as input, offering a compelling alternative to retrieval-based methods. However, this approach comes with significant computational costs, as handling large inputs requires substantial memory resources, particularly on GPUs, which creates a scalability bottleneck~\citep{liu2023scissorhands}. Furthermore, as the context grows, models often struggle to identify the relevant pieces of information buried in all the clutter~\citep{liu2024lost}.

%Compression emerges as a crucial solution in this space, acting as the key enabler for unlocking even larger-scale processing [add cits, couple of most visible papers on compression]. 
To bridge the gap between massive corpora and limited context windows, researchers are developing compression techniques that condense or filter input text~\citep{jha2024characterizing}.
Some of this strategy focuses on optimizing the model’s key-value (KV) cache, ensuring that essential information is retained even within limited contexts.
%By condensing the most relevant information into a compact form, compression can bridge the gap between retrieval-based methods and full-context processing.
Existing approaches fall into two categories: \textit{query-agnostic} compression
~\citep{zhang2023h2o, devoto2024simple, xiao2023efficient, feng2025identify}, and \textit{query-aware} compression, which dynamically optimizes content based on the query during inference~\citep{li2025snapkv, corallo2024finch}. While the latter leads to highly relevant outputs, it is computationally prohibitive, as it requires recompressing the input for each query, making it impractical for real-world deployment.

In this study, we introduce a \textit{task-aware}, query-agnostic compressed cache, offering a balanced trade-off between efficiency and relevance. Instead of recompressing the input for every query, our approach precomputes a compressed cache tailored to a broader task context. 
Our method delivers performance that significantly surpasses existing query-agnostic compression and closely approaches the effectiveness of query-aware compression. 
Figure~\ref{fig:main_image} shows the quality performance of KV cache compression methods against RAG on a 32k token corpus. With high compression rates 64x and 32x (corpus compressed to 512 and 2048 tokens) our method outperforms RAG by about 20 absolute points. 
%our Zero Shot method is comparable to RAG, while our Few Shot method obtains much higher results. 
With compression rates between 16x and 8x, Few Shots performs on par or better than fitting the original corpus in the model context. 
%and gets closer to the results obtained by online compression. % executing compression at inferences time (question aware compression).
%In contrast with the poor results of the corpus-agnostic baseline, our approach retains crucial information for the task at hand. %to achieve quality comparable to or exceeding the execution using the full-context.
%\fabio{At this point, we should introduce the discussion of the first plot (currently Fig. 1), showcasing how our compressed cache retains key information while matching or even outperforming full-context models and baselines.}

Figure~\ref{fig:intro} shows our compression strategy. 
%In this QA example, the documents in the reference corpora require an LLM's KV cache of 128k tokens. 
The task context can be defined through a succinct description (namely Zero Shot) or a small set of representative examples in a Few-Shot setting.
This prompt %process, guided by task instructions and few-shot examples, 
produces a more compact representation of the original context
 while preserving crucial details. 
  Compression happens only once, creating a representation that can be reused for any query within the task domain.
This eliminates the need for repeated processing, streamlining inference by bypassing real-time retrieval and prefilling.
%\fabio{Here, we should include a description of the introductory figure from Giulio, illustrating the core idea of the compressed cache}.
 This approach is not limited to QA but can be applied to a wide range of tasks.

Experimental results across diverse tasks using Llama 3.1, including the LongBench and LongBench v2 benchmarks, demonstrate that our task-aware compression method consistently outperforms both retrieval-based approaches and full-context models in diverse evaluation settings, such as Code Completion and Document QA. Furthermore, experiments on synthetic datasets highlight the superior capability of our method in handling broad, multifaceted queries. Notably, in scenarios requiring the synthesis of widely distributed information, our approach significantly outperforms RAG, establishing compression as a key enabler for scaling LLM reasoning beyond retrieval-based methods.

\section{Background}
\label{sec:knowledge_augmentation}

%\subsection{Transformers' Inference}
LLMs built on the transformer architecture~\citep{NIPS2017_3f5ee243} have become the backbone of modern natural language processing. Given a sequence of $n$ tokens $\mathbf{x} \in \mathbb{R}^n$, each transformer layer produces hidden representations via a multi-head self-attention mechanism:
\begin{align*}
    \mathrm{Attention}(\mathbf{Q}, \mathbf{K}, \mathbf{V})
      &= \mathrm{softmax}\!\Bigl(\frac{\mathbf{Q}\mathbf{K}^\top}{\sqrt{d_k}}\Bigr)\,\mathbf{V},
\end{align*}
where 
\[
\mathbf{Q} = \mathbf{W}^Q \mathbf{h}, \quad
\mathbf{K} = \mathbf{W}^K \mathbf{h}, \quad
\mathbf{V} = \mathbf{W}^V \mathbf{h},
\]
with \(\mathbf{h}\) representing the hidden states (token embeddings) for the input sequence. The dimension \(d_k\) is $\tfrac{d}{H}$ where $d$ is the hidden size and $H$ is the number of attention heads.

In a knowledge-intensive task setup~\citep{petroni2020kilt}, many instruction-tuned LLMs organize their input as a context followed by a user prompt. Formally, let $x$ denote a sequence of tokens. The input sequence can be expressed as:
\begin{equation}
  \label{eq:prompt_context}
  \mathbf{x} \;=\;
  \Bigl[\;\mathbf{x}^{(\text{cont})},\;\mathbf{x}^{(\text{prompt})}\Bigr] \in \mathbb{R}^{n^{(\text{cont})} + n^{(\text{prompt})}},
\end{equation}

Crucially,
\(
\mathbf{x}^{(\text{cont})} \)
serves as the \emph{knowledge} the model has access to when generating the final response.
%A popular strategy to supply LLMs with knowledge is RAG. RAG uses an external retriever to select only the most relevant chunks. Those retrieved chunks then form the model’s context:
%\[
%\mathbf{x}^{(\text{cont})} 
%\;\; \gets\;\;
%\text{Retriever}(\text{query}=\mathbf{x}^{(\text{prompt})}).
%\]
%\subsection{Retrieval-Augmented Generation (RAG)}
%A popular strategy to supply LLMs with knowledge is RAG. RAG uses an external retriever to select only the most relevant chunks. Those retrieved chunks then form the model’s context:
%\[
%\mathbf{x}^{(\text{cont})} 
%\;\; \gets\;\;
%\text{Retriever}(\text{query}=\mathbf{x}^{(\text{prompt})}).
%\]
%While RAG has proven highly effective for narrow queries that rely on a few evidence pieces, it struggles with broader queries that require reasoning over long documents, often missing crucial information and reducing response accuracy.
%\subsection{Long-Context Models}

%Recent LLMs have significantly expanded their context windows to hundreds of thousands of tokens, enabling direct processing of large corpora without explicit retrieval. Instead of selecting a small subset of relevant passages via retrieval mechanisms, one could place the entire corpus within \(\mathbf{x}^{(\text{cont})}\), allowing the model to access and reason over a much larger body of information. However, these models still face fundamental constraints: (i) finite context sizes limit the amount of information that can be stored in a single pass, and (ii) self-attention over long sequences incurs substantial memory overhead, making inference impractical at scale.

During inference, an LLM operates in two distinct phases:

\paragraph{Prefill Stage.} The model processes the entire input sequence $\mathbf{x}$ and caches the key-value (KV) matrices for each layer:
\begin{equation}
  \mathbf{K} \in \mathbb{R}^{n \times d}, \quad \mathbf{V} \in \mathbb{R}^{n \times d}.
\end{equation}

\paragraph{Generation Stage.} Tokens are generated autoregressively. For each new token $y_j$, the model computes:
\begin{equation}
  \mathbf{q}^{\text{new}}, \mathbf{k}^{\text{new}}, \mathbf{v}^{\text{new}} \in \mathbb{R}^{d},
\end{equation}
and updates the cache as follows:
\begin{equation}
  \mathbf{K} \leftarrow \begin{bmatrix} \mathbf{K} \\ \mathbf{k}^{\text{new}} \end{bmatrix}, \quad
  \mathbf{V} \leftarrow \begin{bmatrix} \mathbf{V} \\ \mathbf{v}^{\text{new}} \end{bmatrix}.
\end{equation}
Thanks to the cached KV matrices, self-attention complexity reduces from $O(n^2 d)$ to $O(nd)$, significantly improving efficiency. However, for large $n^{(\text{cont})}$, storing these matrices for every layer can be prohibitively memory-intensive.

\subsection{KV-Cache Compression}
To mitigate the memory load from very long contexts, a promising approach is KV-cache compression. Instead of retaining \(\mathbf{K},\mathbf{V}\) for all $n$ tokens, one compresses them into smaller matrices:
\[
\widetilde{\mathbf{K}} \in \mathbb{R}^{k\times d}
\quad\text{and}\quad
\widetilde{\mathbf{V}} \in \mathbb{R}^{k\times d}
\quad\text{with}\quad k \ll n,
\]
that still preserve the essential information needed for generating the final response. Formally, one seeks to minimize a divergence measure,
\[
  \min_{\widetilde{\mathbf{K}},\,\widetilde{\mathbf{V}}}
  \Bigl[\,
    \mathrm{dist}\bigl(\mathbf{y}\!\mid\!\mathbf{K},\mathbf{V},\;\mathbf{y}\!\mid\!\widetilde{\mathbf{K}},\widetilde{\mathbf{V}}\bigr)
  \Bigr],
\]
where \(\mathbf{y}\) is the model’s output.

%\paragraph{KV Compression for Knowledge Intensive Tasks}
Previous work has also recognized that compressing the KV cache of LMs allows for improved performance at much smaller memory costs \citep{qin2023dodo,ge2023model,corallo2024finch}. In general there are three levels of compression: (1) task-agnostic compression where no task information is present \citep{charikv,zhang2024lorc}, (2) ad-hoc compression, where the compression is tailored for a single specific task such as question-answering \citep{fu2024not,cai2024pyramidkv}, and (3) query-aware compression where the compression happens w.r.t. a specific example \citep{rehg2024kv,xu2024think}. Our work proposes a new compression technique that works for \textit{any task} specified in the prompt.

To explain how compression works, we detail a query-aware iterative approach that compresses the KV cache 
by retaining only the most relevant Key-Value vectors for the query given at inference time~\citep{corallo2024finch}. 
Let \(m\) be the chunk size, and let \(\{\mathbf{c}_1, \mathbf{c}_2, \dots\}\) 
be the segments obtained by slicing the context \(\mathbf{x}^{(\text{cont})}\). 
The segment-to-document ratio is defined as
\begin{equation}
    s = \frac{n}{m}
    \label{eq:segment_ratio}
\end{equation}
where \( n \) is the total document length.
At iteration~\(i\), the method takes as input
\[
\Bigl[
   \underbrace{\widetilde{\mathbf{K}}_{i-1}, \widetilde{\mathbf{V}}_{i-1}}_{\text{previous compressed cache}},\;
   \underbrace{\mathbf{c}_i}_{\text{current chunk}},\;
   \underbrace{\mathbf{q}}_{\text{question}}
\Bigr],
\]
where
\(
\widetilde{\mathbf{K}}_{i-1}, \widetilde{\mathbf{V}}_{i-1} \in \mathbb{R}^{r_{i-1}\times d}
\)
denote the compressed cache from the previous iteration,
\(
\mathbf{c}_i \in \mathbb{R}^{m\times d}
\)
is the chunk of context for the current iteration,
and
\(
\mathbf{q} \in \mathbb{R}^{q \times d}
\)
is the question to be answered.

During the forward pass, the multi-head attention scores are computed. 
Crucially, the \emph{cross-attention} submatrix
\[
\mathbf{W}^{(\mathbf{q},\mathbf{c})} \;\in\; \mathbb{R}^{q \times (r_{i-1} + m)},
\]
captures how each question token attends to both the previous cache and the current chunk. 
%\textsc{Finch} 
The online method then selects the top \(r\) token positions (according to the highest attention scores in 
\(\mathbf{W}^{(\mathbf{q},\mathbf{c})}\)) to form
\(
\widetilde{\mathbf{K}}_{i}, \widetilde{\mathbf{V}}_{i}.
\)
After processing all chunks, the final 
\(\widetilde{\mathbf{K}}, \widetilde{\mathbf{V}} \in \mathbb{R}^{k \times d}\) 
provide a global representation of the entire context, at a substantially reduced size. Agnostic methods use similar principles but in a single offline computation of the cache, thus without making use of the query.

\subsection{RAG and Knowledge Intensive Tasks}
Knowledge-intensive tasks require models to utilize external information to arrive at accurate answers. This category encompasses tasks like question-answering, summarization, and fact-checking \citep{kwiatkowski2019natural,petroni2020kilt}. While larger models have demonstrated improved capacity to store knowledge whitin their parameters \citep{tirumala2022memorization,biderman2023emergent,wang2024generalization}, they face limitations, such as difficulties in updating or modifying the embedded information \cite{de2021editing} and hallucinations \cite{ji2023survey}. 
To address this, RAG integrates retrieval mechanisms with generative language models, enhancing the accuracy by explicitely incorporating external knowledge~\citep{lewis2020retrieval,borgeaud2022improving,gao2023retrieval}. However, RAG has its own challenges: crucial information might not be included in the top-ranked retrieval results, preventing the language model from reasoning over it.

\section{Problem Formulation}
A fundamental challenge in compressing long-context representations is achieving \textit{query-agnostic} compression, where a precomputed compressed cache 
 \(\widetilde{\mathbf{K}}, \widetilde{\mathbf{V}}\)
  remains effective for any query at inference time. However, empirical results indicate that existing query-agnostic methods exhibit significant performance degradation, particularly under high compression rates, often falling behind both full-context processing and RAG~\citep{kvpress}.

On the other hand, query-aware compression, circumvents the challenges of long-context models by (i) processing documents in smaller segments while adjusting positional embeddings accordingly, and (ii) reducing the KV memory footprint in proportion to the compression rate. In practice, query-aware compression can outperform retrieval-based methods like RAG. However, a critical drawback is its reliance on a \textit{specific} user query at compression time. While effective for single-query scenarios, this assumption becomes impractical when multiple queries need to be processed, as rerunning the compressor for each query is computationally prohibitive, thus undermining the original goal of avoiding large-scale retrieval or excessive context expansion.

This raises a key research question: \textbf{Can we design a query-agnostic compression method that preserves efficiency while maintaining qualitative competitive performance?}

\begin{figure*}[ht]
    \centering
    \includegraphics[width=\textwidth]{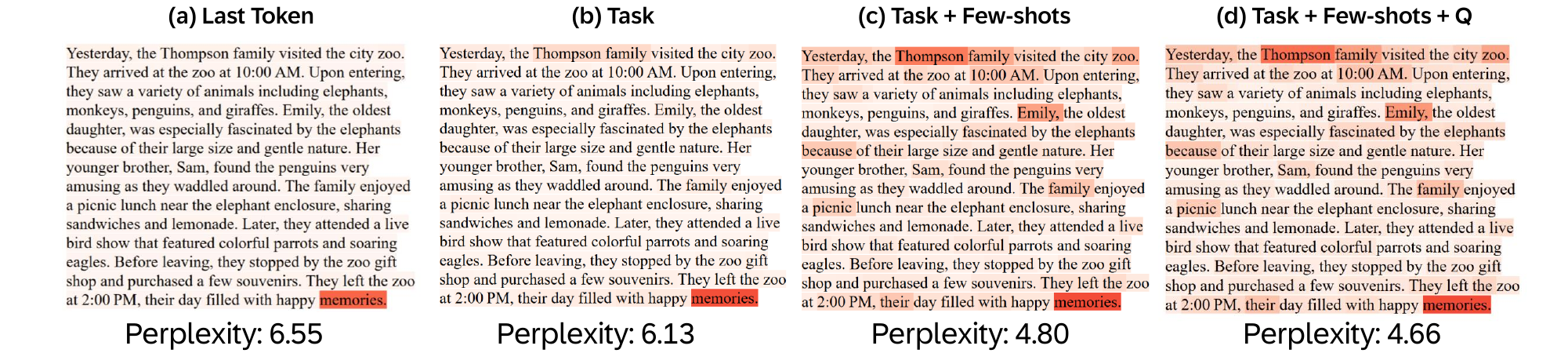}
    \caption{We examine how the model attends to context tokens when conditioned on the last token, a task description, a description with few-shot examples, and a description with both few-shot examples and a question. As we increase the information in the prompt, %from only last token, to task description, to few-shot examples, to providing the question 
    the cross attention between the prompt and the context better discriminates the context tokens that are relevant for decoding the answer. The perplexity is calculated on the loss for the answer.}
    \label{fig:method}
\end{figure*}

\section{Methodology}
\label{sec:methodology}
In this section, we present our \textit{task-aware, query-agnostic} compression strategy, motivated by the remarkable in-context learning capabilities of modern LLMs. We describe how we obtain %adapt  \textsc{Finch} into 
a single, reusable cache that covers an entire task's external knowledge.

\subsection{Motivation via In-Context Learning}
LLMs demonstrate remarkable in-context learning capabilities~\citep{brown2020language,dong2022survey}: once they read a sufficiently large prefix of tokens, they often infer the nature of the task without any parameter updates. In practice, tokens later in the input are easier to predict because the model has more context to condition on. An intuitive analogy is a student preparing for exams.

When asked a new question with no references at hand (i.e., \emph{zero-shot inference}), both the human student and the LLM rely solely on prior knowledge. If a few solved examples are provided (\emph{few-shot learning}), they adapt by identifying solution patterns from the examples, thereby reducing uncertainty. Giving the student all available reference material (akin to letting an LLM observe the entire corpus) can maximize accuracy.
Our objective is to construct a \emph{compressed representation} of the corpus in advance, much like a curated ``cheat sheet'' that condenses the key information. 
To explore how LLMs dynamically allocate attention within their input, we analyze cross-attention patterns across different prompting strategies, as shown in Figure~\ref{fig:method}. Notably, in (c), when a task description and few examples are used, the model commits to a subset of tokens in the latent space that is similar to (d), the query-aware case, as evidenced by the qualitative similarity in attention distribution and the corresponding reduction in perplexity on the final answer. This suggests that a sufficiently structured prompt allows the model to internally resolve much of the task-specific ambiguity, even before an explicit question is introduced.
%This phenomenon is reminiscent of the progressive commitment to latent abstractions observed in Diffusion Models. In particular, as shown in prior work on hierarchical feature emergence in generative processes~\citep{franzese2024latent}.
Guided by these insights, our goal is to construct a \textit{task-aware, query-agnostic} compression strategy that enables efficient, reusable caching of task-relevant knowledge.

\subsection{Task-Aware, Query-Agnostic Compression}
Concretely, we create a compressed cache capable of supporting any query within a defined task domain. The procedure is:

\textbf{(1) Define a Task Description ($\mathbf{T}$).}
Rather than targeting a single query, we incorporate a Task Description specifying the type of questions to be answered (e.g.,~“Answer factual questions about this corpus”). When available, we add a few examples to better illustrate the target task. % desired style and level of detail.

\textbf{(2) Offline Compression.}
We adapt an existing query-aware method by replacing the query with the Task Description $\mathbf{T}$~\citep{corallo2024finch}. The corpus is split into chunks $\{\mathbf{c}_1, \mathbf{c}_2, \dots\}$, and we iteratively compress these chunks together with $\mathbf{T}$:
\[
\bigl[\widetilde{\mathbf{K}}_{i-1}, \widetilde{\mathbf{V}}_{i-1}, \mathbf{c}_i, \mathbf{T}\bigr].
\]
This yields a compressed cache $(\widetilde{\mathbf{K}}, \widetilde{\mathbf{V}})$ that captures essential domain knowledge. Crucially, this compression is computed only once.

\textbf{(3) Efficient Query Resolution.}
When facing a new query $\mathbf{q}_{\mathrm{new}}$ from the same domain, we prepend the precomputed cache:
\[
\mathbf{x}^{(\text{prompt})} \;=\; \bigl[\widetilde{\mathbf{K}}, \widetilde{\mathbf{V}}, \mathbf{q}_{\mathrm{new}}\bigr].
\]
No further compression or retrieval is necessary. The LLM conditions on $\widetilde{\mathbf{K}}$ and $\widetilde{\mathbf{V}}$ to answer, reusing the relevant external knowledge.

We develop two variants of this approach: \textbf{KVCompress ZS (Zero-Shot Task Description)} uses only broad task instructions, and \textbf{KVCompress FS (Few-Shot Task Description)} includes sample task examples (such as QA pairs). 
Empirically, this \emph{offline, query-agnostic compression} speeds up inference while offering more robust coverage of multi-hop or broad queries than RAG’s similarity-based lookups. In contexts where RAG struggles to gather widely dispersed evidence or gets confused by near-identical entity names, the global coverage of the compressed cache avoids such pitfalls.

\section{Experimental Setup}
\label{sec:experimental_setup}
\subsection{Models}
Our experiments use \textsc{Llama-3.1-8B-Instruct}~\citep{dubey2024llama} with 5 knowledge-infusion baselines:
\textsc{RAG}, \textsc{KVCompress~ZS}, \textsc{KVCompress~FS}, \textsc{KVCompress~FS+Q}, and a Full Context upper bound.
For the compression baselines, we set \( s = 2 \) in all experiments, as defined in Equation~\eqref{eq:segment_ratio}.

For RAG, we use \textsc{bge-large-en-v1.5}~\citep{bge_embedding} as the retriever, filling the entire context with the top-$k$ retrieved documents, each containing 256 tokens. The same prompt is used across all models.\footnote{Detailed prompt configurations for each dataset are available in our configuration files: \url{https://anonymous.4open.science/r/context-compression-2-6B58/conf/custom_datasets/}.}

\subsection{Datasets}
We evaluate these methods on three datasets:

\paragraph{LongBench v2}~\citep{bai2024longbench} is a multiple-choice benchmark designed to evaluate the reasoning capabilities of LLMs on realistic long-context multitasks. It comprises questions distributed across six major categories: single-document QA, multi-document QA, long in-context learning, long-dialogue history understanding, code repository understanding, and long structured data understanding. Context lengths vary significantly, from 8,000 to 2 million words. State-of-the-art models, without enhanced reasoning prompts, attained only 50.1\% accuracy.
In our experiments, we evaluate using Exact Match scores. The ground truth for evaluation consists of selecting one option among A, B, C, or D. For the full context evaluation, we use the entire context, truncating to keep the first and last halves if it exceeds the context window length, and constrained generation over the provided options with 1 $\text{max\_new\_tokens}$~\citep{de2020autoregressive}.

\paragraph{LongBench}~\citep{bai2023longbench} is a benchmark also designed for long-context understanding. It covers 21 datasets across six tasks, including single-document and multi-document QA, summarization, code completion, few-shot learning, and a synthetic task. The benchmark has an average context length of 6,711 words for English tasks. LongBench uses automated evaluation metrics, such as F1 for question answering tasks~\citep{rajpurkar2016squad} and ROUGE-L for summarization tasks~\citep{lin-2004-rouge}, and Edit Similarity for the Code Completion task~\citep{svyatkovskiy2020intellicode}.

\paragraph{Our Synthetic Dataset}
is designed %to better understand the strengths of KV-cache compression methods, we design synthetic QA datasets 
to enable full control over the interconnectivity among text chunks in the corpus for a QA task. We use two query types: \textit{direct retrieval}, which retrieval systems can handle easily, and \textit{join-like queries}, requiring broader comprehension of the entire corpus. Ground truth answers are generated as lists of entities so that model evaluation is straightforward: predictions and ground truths are normalized (removing punctuation and converting to lowercase) and a score is obtained by computing word overlap between the predicted and ground truth entities. We describe the dataset creation next.

\begin{figure}[t]
    \centering
    \includegraphics[width=\columnwidth]{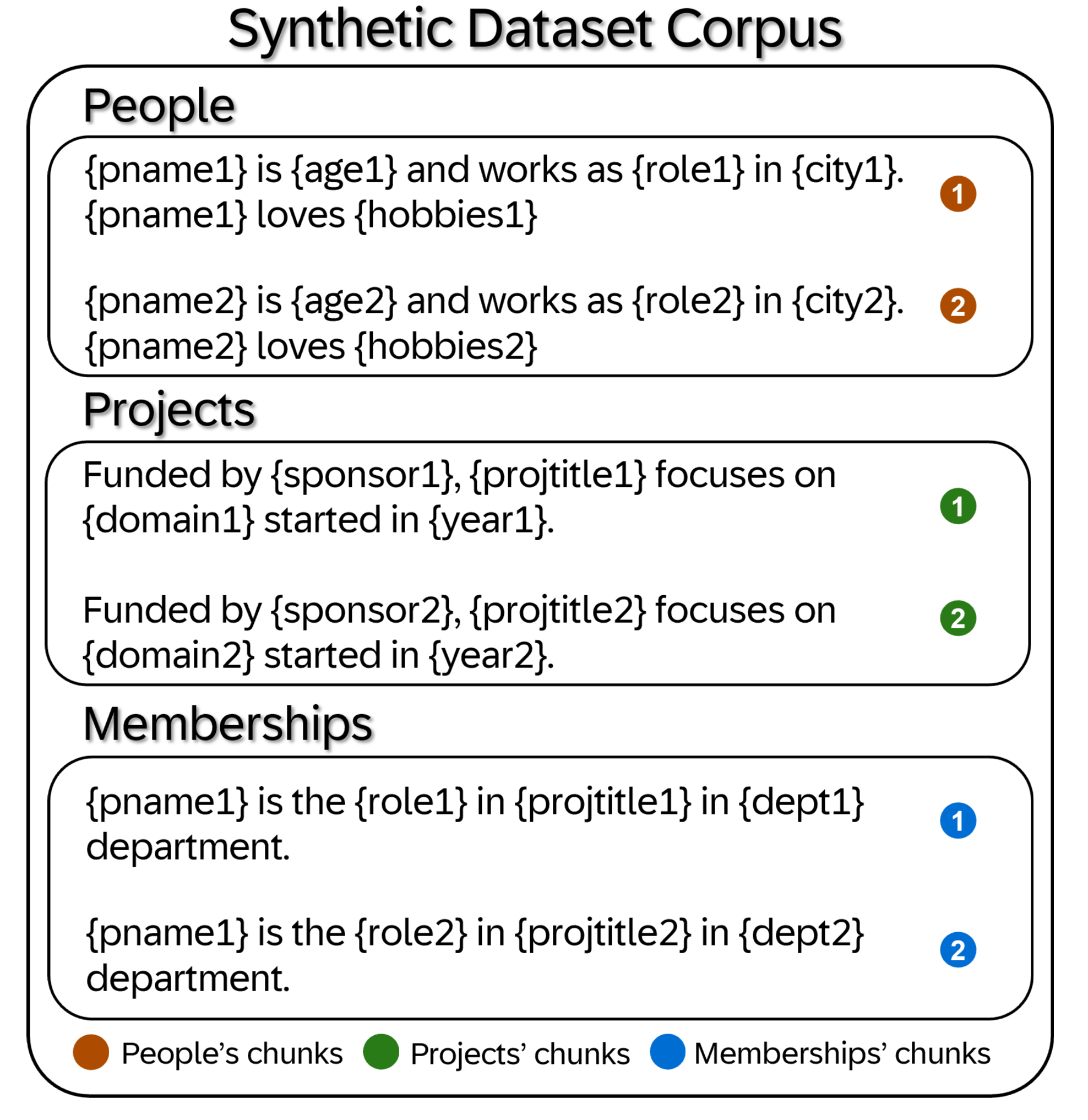}
    \caption{Overview of our synthetic dataset. In this example, the connectivity level is set to 2. }
    \label{fig:dataset_overview}
\end{figure}
\begin{figure}[t]
    \centering
    \includegraphics[width=\columnwidth]{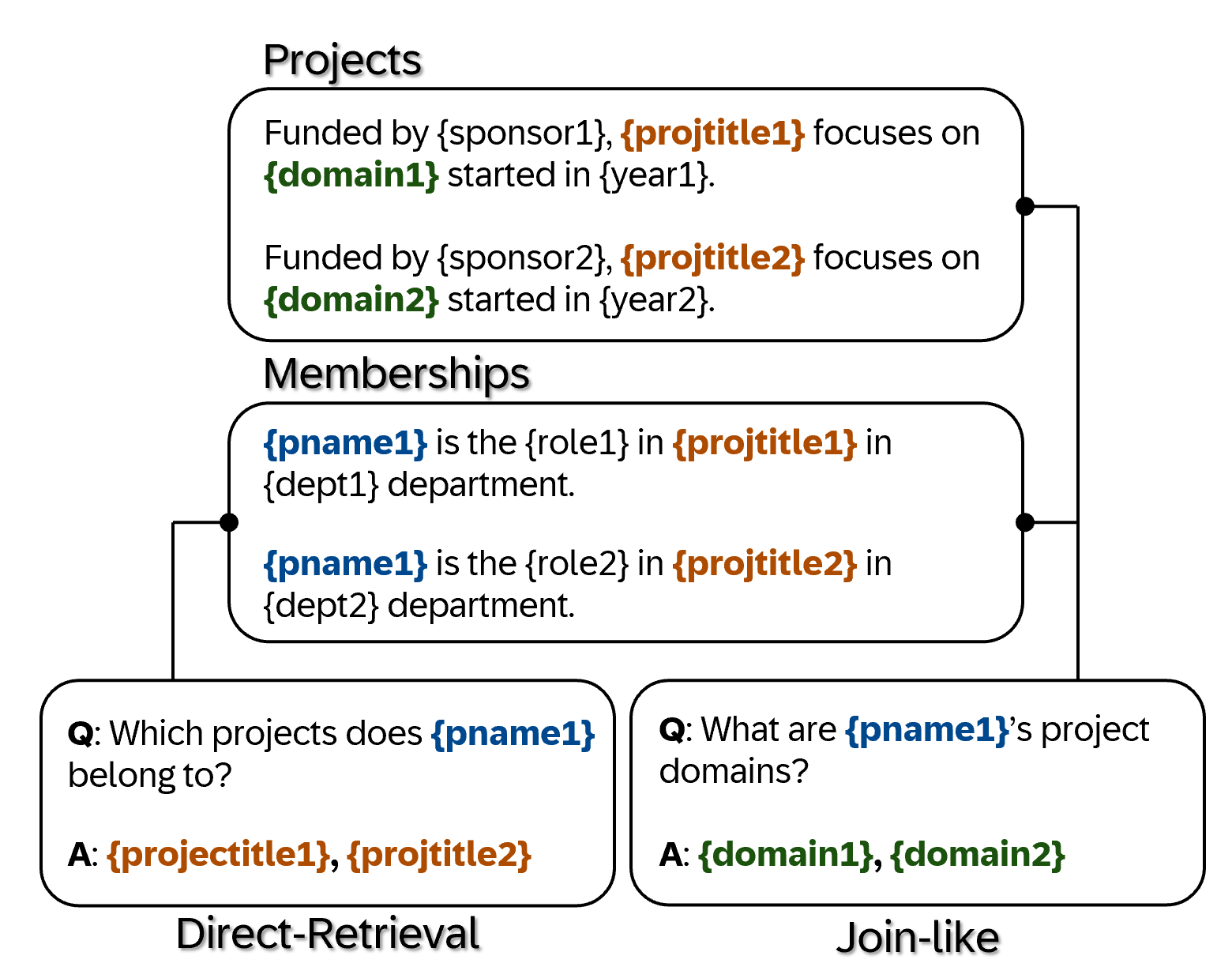}
    \caption{Overview of our questions. In this example, the connectivity level is set to 2.}
    \label{fig:dataset_questions}
\end{figure}

\section{Synthetic Dataset Construction}
\label{sec:synthetic_data_construction}
%To systematically evaluate KV-cache compression methods and compare them against retrieval-based approaches, 
We construct a synthetic dataset designed to precisely control corpus complexity and the connectivity level between text chunks. By varying inter-chunk connectivity, we are able to thoroughly evaluate different methods, identifying the exact scenarios where each technique performs well or fails. %Our findings offer valuable insights that can help advance the state-of-the-art in query-agnostic compression strategies. 
Figure~\ref{fig:dataset_overview} illustrates the structured design of our corpus. Our dataset will be publicly released to support future research. % and facilitate further advancements in this area. %\orion{Do we want to name this dataset?}

\subsection{Structured Entities and Corpus Chunks}
We define three entity types. % organized into text chunks.

\begin{figure*}[t]
    \centering
    \includegraphics[width=\textwidth]{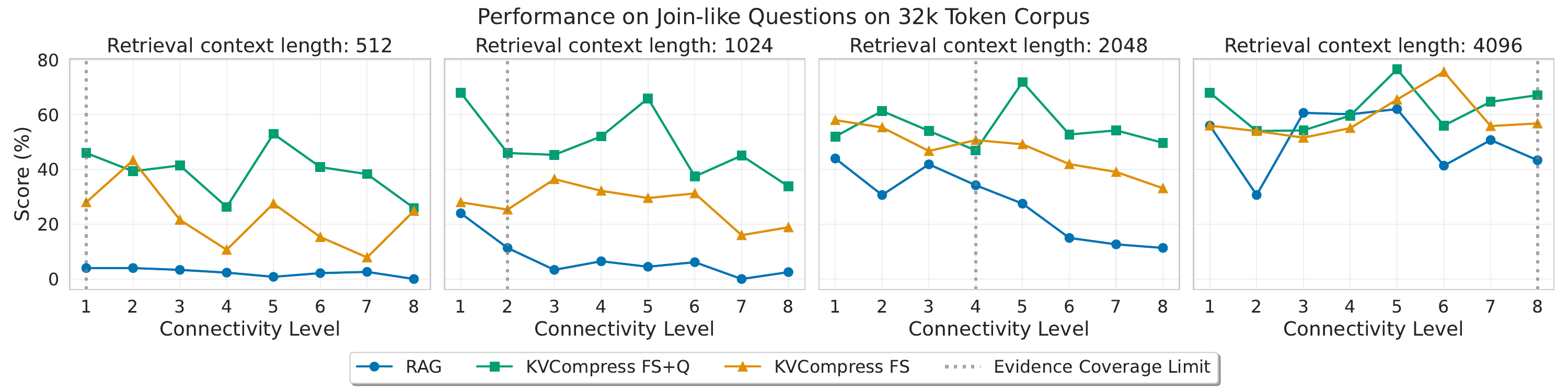}
    \caption{Performance by increasing target context length (64x to 8x compression rate) and connectivity level for Join-like questions in the synthetic dataset. The dashed line 
    indicates for which connectivity level RAG gets the needed chunk for a given context length.} % can cover up to with evidences indicates the level's chunk that fit the context in RAG.}
    \label{fig:hard_questions}
\end{figure*}

\stitle{People.} Each person is described through template-structured biographies containing attributes such as \textit{name}, \textit{age}, \textit{occupation}, \textit{city}, and \textit{hobbies}. To maintain uniformity and facilitate controlled experiments, each biography text chunk is standardized to a length of 256 tokens using additional neutral filler text.

\stitle{Projects.} Each project has attributes including \textit{title}, \textit{domain}, \textit{sponsor}, \textit{year started}, and a descriptive summary. Like for people, each text chunk is standardized to 256 tokens.

\stitle{Memberships.} A membership  represents the relationship between people and projects and specifies the \textit{role} (e.g., \textit{Engineer}, \textit{Manager}) and \textit{department} (e.g., \textit{R\&D}, \textit{Marketing}) that a person holds in a project. These text chunks similarly include filler text to meet the fixed-length criterion.

We generate multiple corpus instances with varying \emph{connectivity levels}, ranging from 1 to 8, where level $k$ means each person links to exactly $k$ projects. Higher connectivity increases dataset complexity by distributing relevant information about a person across multiple membership and project chunks, thus challenging the model's ability to synthesize scattered information. Each corpus at a given connectivity level comprises exactly 32k tokens, ensuring consistent corpus size across experiments.

\subsection{Controlled Question Types for Evaluation}
To rigorously evaluate the performance of both KV-cache compression and retrieval-based methods, we generate two primary question categories (Figure~\ref{fig:dataset_questions}).

\paragraph{Direct Retrieval Questions.} These questions require information localized within a single \textit{Memberships} chunk. Example templates include:
\setlist{nolistsep}
\begin{itemize}[noitemsep]
\item \textit{Which projects does {\texttt{pname}} belong to?}
\item \textit{Which role does {\texttt{pname}} have in {\texttt{projtitle}}?}
\item \textit{Which department is {\texttt{pname}} part of?}
\end{itemize}
Where variables like \texttt{pname} are instantiated at generation time. Answering these queries does not require cross-chunk synthesis.

\paragraph{Join-like Questions.} Answering these queries require combining information across multiple \textit{Memberships} and \textit{Projects} chunks. For example:
\setlist{nolistsep}
\begin{itemize}[noitemsep]
\item \textit{What are {\texttt{pname}}'s project domains?}
\item \textit{In which years did {\texttt{pname}}'s projects begin?}
\item \textit{Who sponsors {\texttt{pname}}'s projects?}
\end{itemize}

Addressing these queries tests a model's capability for multi-hop reasoning and synthesis across distributed knowledge sources. As the connectivity level grows, these join-like questions become increasingly complex, requiring the aggregation of information from multiple chunks.

For each connectivity level (1 through 8), we generate 50 distinct queries: 25 direct-retrieval and 25 join-like, totaling 400 distinct queries across all connectivity levels. Additionally, we create a targeted variant of direct retrieval questions with highly similar entity names (e.g. Person\_01, Person\_02, \dots) to test embedding-based retrieval robustness against closely related entities.

\section{Results and Discussions}

We structure our analysis around five research questions over our synthetic dataset experiments while also drawing connections to the LongBench and LongBench v2 results.

\stitle{When Does RAG Fail?} RAG exhibits significant limitations in multi-hop reasoning and high-connectivity scenarios. Our synthetic dataset reveals a sharp performance decline for join-like questions — those requiring the integration of dispersed information (Figure~\ref{fig:hard_questions}). This degradation stems from RAG’s reliance on similarity-based retrieval and frequently omits crucial chunks containing necessary details, thus limiting the model’s ability to accurately answer these queries. 

\begin{figure}[t]
    \centering
    \includegraphics[width=\columnwidth]{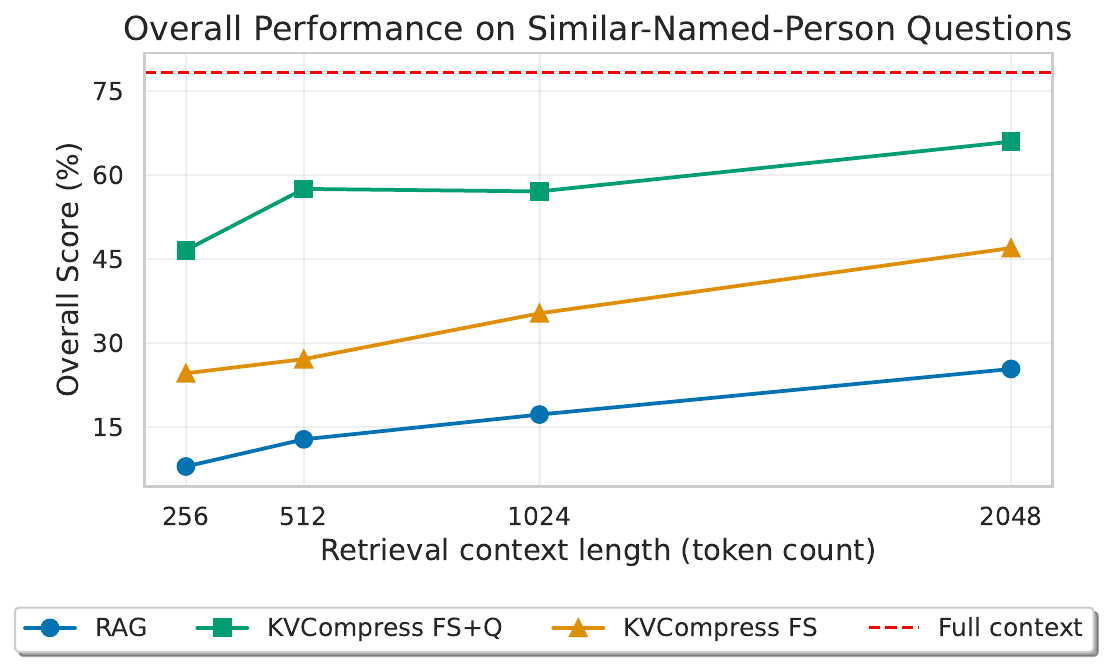}
    \caption{Performance by retrieval context length size (128x to 16x compression rate)  for Direct Retrieval questions with highly similar entity names in the synthetic
    dataset.} % Dashed line indicates the results using an LLM with a 32k context size.}
\label{fig:cloze_easy_question}
\end{figure}

\begin{figure}[t]
    \centering
    \includegraphics[width=\columnwidth]{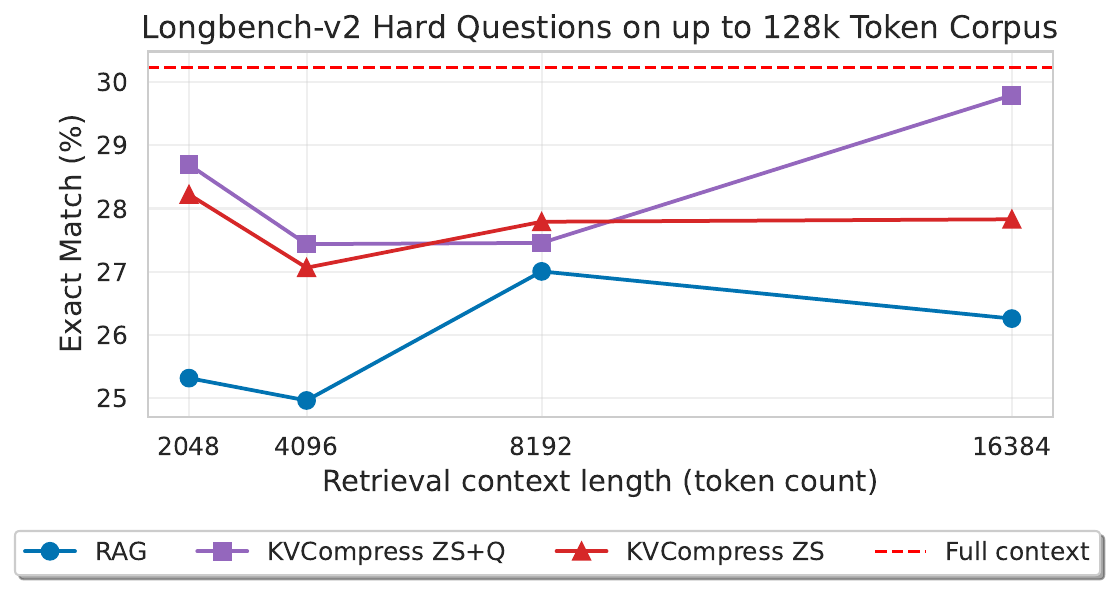}
  \caption{Performance by retrieval context length size (64x to 8x compression rate)  for the Hard Questions in
    LongBench v2.}
  \label{fig:longbench_v2}
\end{figure}

Additionally, RAG struggles with entity disambiguation, particularly for similarly named entities (Figure~\ref{fig:cloze_easy_question}). Embedding-based retrieval frequently misidentifies relevant passages and retrieves irrelevant chunks due to embedding proximity among similarly named entities. These limitations are further reflected in the results for the LongBench v2 ``hard questions'' (Figure~\ref{fig:longbench_v2}), where \textsc{KVCompress ZS} already
%\textsc{KVCompress ZS+Q}
outperforms RAG, despite the absence of few-shot examples. RAG gains only marginal improvements with longer contexts, highlighting the limitations of retrieval-based methods in scenarios that require complex long-context understanding. 

\begin{figure*}[t]
    \centering
    \includegraphics[width=\textwidth]{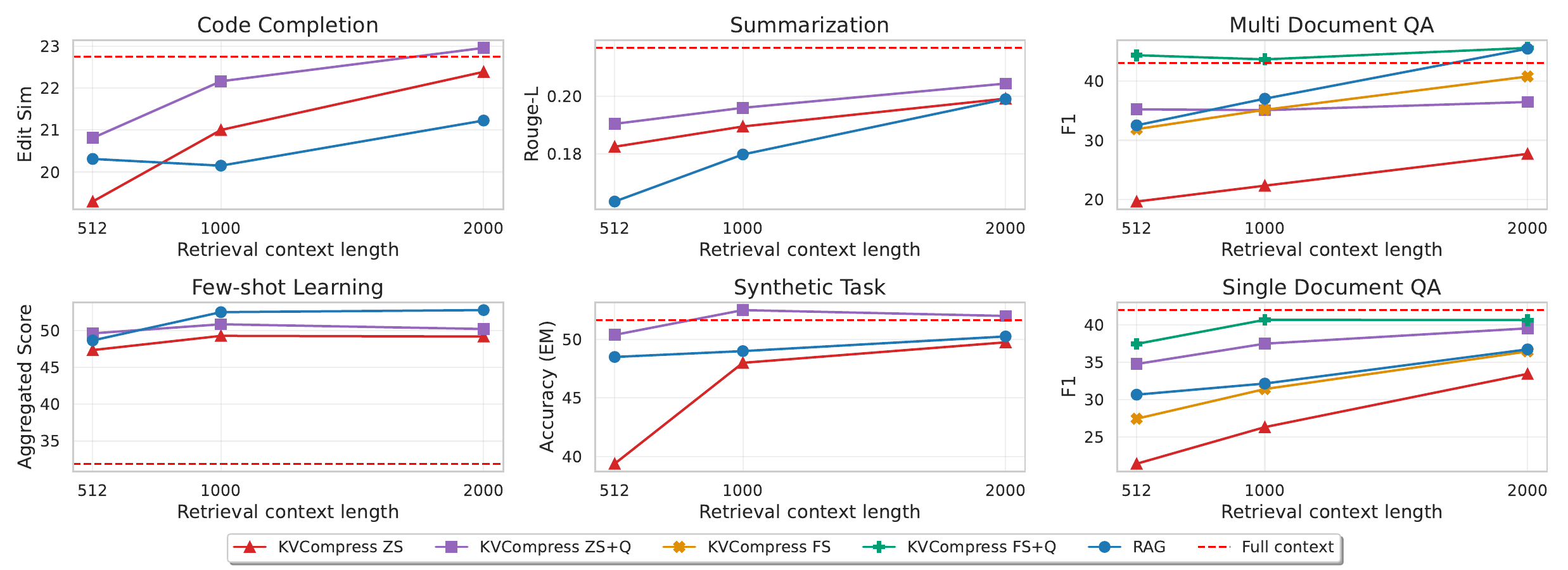}
    \caption{Performance results on Longbench. Our FS variant is reported when examples are available (QA tasks). For the QA tasks, the examples used in KVCompress FSt are taken (and removed) at random from the test set.}
    \label{fig:longbench}
\end{figure*}

In LongBench (Figure~\ref{fig:longbench}), \textsc{KVCompress ZS} consistently surpasses RAG in summarization and code completion tasks, which require synthesizing multiple passages or maintaining a broader context and is competitive in the other tasks, except the QA, where it is the Few Shot setting that competes with RAG.

\begin{tcolorbox}[colframe=black, colback=gray!10, sharp corners=south, boxrule=0.1pt]\small{\textbf{Takeaway:} RAG fails in multi-hop reasoning and entity disambiguation, limiting its effectiveness in synthesis tasks with long-context.}\end{tcolorbox}

\begin{figure}[t]
    \centering
    \includegraphics[width=\columnwidth]{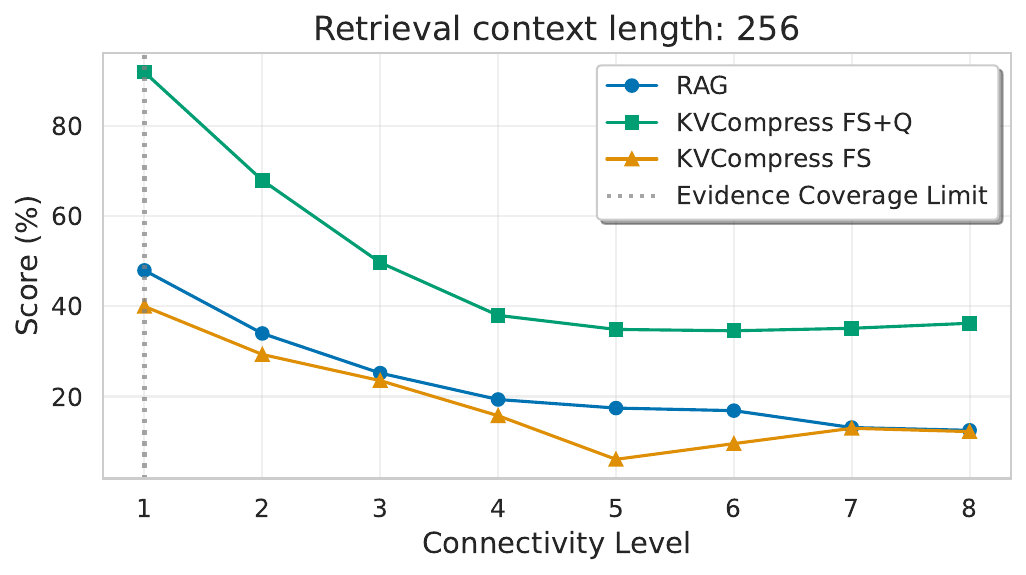}
    \caption{Performance by connectivity level (64x compression rate)  for the Direct Retrieval questions in the synthetic dataset.}
    \label{fig:easy_question_256}
\end{figure}

\stitle{When Is RAG Effective?} Experiments show that RAG performs well when answers are localized within a single document chunk. In direct-retrieval scenarios, where each answer is self-contained, RAG’s %similarity-based retrieval 
proves effective. This trend is evident in our synthetic dataset (Figure~\ref{fig:easy_question_256}) and in LongBench results (Figure~\ref{fig:longbench}), where RAG achieves competitive performance on tasks such as direct question answering.

\begin{tcolorbox}[colframe=black, colback=gray!10, sharp corners=south, boxrule=0.1pt]{\small \textbf{Takeaway:} RAG excels when answers are self-contained within the retrieved chunks, making it effective for narrowly scoped questions.}\end{tcolorbox}

\begin{table}[t]
  \centering
  \begin{adjustbox}{max width=\columnwidth,center}
    \begin{tabular}{llcc}
      \toprule
      \textbf{Dataset (Metric)} & \textbf{R. Length} & \textbf{KVCompress ZS} & \textbf{RAG} \\
      \midrule
      \multirow{4}{*}{\shortstack[l]{GovReport \\ (Rouge-L $\uparrow$)}}
        & Full context  & \multicolumn{2}{c}{24.63} \\ \cmidrule(lr){2-4}
        & 512    & \textbf{21.03} & 19.14 \\
        & 1000   & \textbf{21.57} & 19.92 \\
        & 2000   & \textbf{22.58} & 21.59 \\
      \midrule
      \multirow{4}{*}{\shortstack[l]{MultiNews \\ (Rouge-L $\uparrow$)}}
        & Full context   & \multicolumn{2}{c}{18.45} \\ \cmidrule(lr){2-4}
        & 512    & \textbf{16.79} & 9.35 \\
        & 1000   & \textbf{17.86} & 13.10 \\
        & 2000   & \textbf{18.42} & 17.39 \\
      \midrule
      \multirow{4}{*}{\shortstack[l]{PassageCount \\ (Accuracy $\uparrow$)}}
        & Full context   & \multicolumn{2}{c}{3.25} \\ \cmidrule(lr){2-4}
        & 512    & \textbf{5.77}  & 1.50 \\
        & 1000   & \textbf{5.50}  & 2.50 \\
        & 2000   & \textbf{5.00}  & 4.00 \\
      \midrule
      \multirow{4}{*}{\shortstack[l]{LCC \\ (Edit Sim  $\uparrow$)}}
        & Full context   & \multicolumn{2}{c}{20.13} \\ \cmidrule(lr){2-4}
        & 512    & 16.62 & \textbf{19.36} \\
        & 1000   & \textbf{19.35} & 17.89 \\
        & 2000   & \textbf{20.66} & 18.61 \\
      \bottomrule
    \end{tabular}
  \end{adjustbox}
  \caption{Performance of KVCompress ZS and RAG across all query-agnostic datasets in LongBench.}
  \label{tab:results}
\end{table}

\stitle{What Are the Benefits of Our Task-Aware, Query-Agnostic Compression Method?} Our proposed method departs from traditional retrieval and query-aware compression approaches by precomputing a unified, task-aware compressed cache that is reusable across multiple queries. This approach offers several advantages over both full-context processing and RAG: it dramatically reduces memory overhead and inference time while preserving a global representation of the corpus. Moreover, it overcomes RAG's limitations in broad query scenarios as shown in Figure~\ref{fig:hard_questions} and in the LongBench v2 results both in Figure~\ref{fig:longbench} and in Table \ref{tab:results}.

\begin{figure}[t]
    \centering
    \includegraphics[width=\columnwidth]{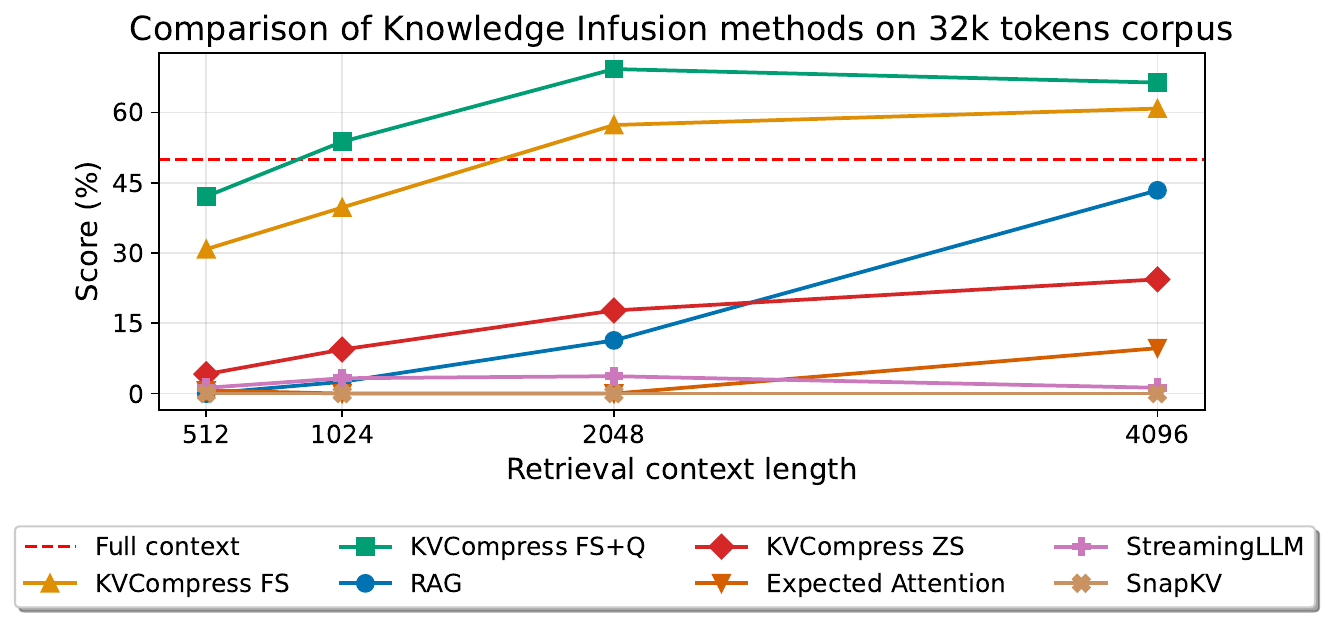}
    \caption{Comparison against query-agnostic compressors on our synthetic dataset for Join-like queries with connectivity level=8 across decreasing compression levels (64x to 8x).}
    \label{fig:competitors}
\end{figure}

Besides RAG, we evaluate our task-aware compression strategies against several baseline compression methods, including Expected Attention~\citep{kvpress}, StreamingLLM~\citep{xiao2023efficient}, and SnapKV (question-agnostic)~\citep{li2025snapkv}.
As shown in Figure~\ref{fig:competitors}, our synthetic dataset — configured with a high-connectivity level of 8 to test multi-hop reasoning — reveals a critical gap in the effectiveness of these  compression techniques. These methods, which treat the context in isolation and are “blind” to the specific requirements of the query, consistently underperform. 
Moreover, the results underscore the utility and sensitivity of our synthetic dataset, which effectively captures nuances in method performance, clearly differentiating between naive and more sophisticated compression techniques. This dataset serves as a robust benchmarking tool for future research into KV-cache compression methods, particularly for evaluating their comprehensive knowledge reasoning %multi-hop reasoning and synthesis 
capabilities 
in challenging scenarios.

\begin{tcolorbox}[colframe=black, colback=gray!10, sharp corners=south, boxrule=0.1pt] \small{\textbf{Takeaway:} Task-aware query-agnostic compression is scalable and efficient achieving better performance when RAG fails. }\end{tcolorbox}

\stitle{How to Choose the Compression Variant?} A key decision is whether to use the zero-shot (ZS) or few-shot (FS) variant. Our results suggest:

%\setlist{nolistsep}
%\begin{itemize} [noitemsep]
%\item 
\textsc{KVCompress ZS} relies solely on a broad task instruction—is optimal for query-agnostic tasks, % like GovReport, MultiNews, PassageCount, and LCC, 
as shown in Table~\ref{tab:results}. In LongBench (Figure~\ref{fig:longbench}), for non-QA tasks, it performs comparably or better than RAG; in hard settings like LongBench v2 (Figure~\ref{fig:longbench_v2}) it outperforms RAG.

\textsc{KVCompress FS} incorporates a few examples and excels in tasks such as question-answering. As shown in Figures~\ref{fig:main_image} and~\ref{fig:longbench}, FS significantly reduces the performance gap with RAG, whereas ZS struggles in these scenarios.

%\end{itemize}

\begin{tcolorbox}[colframe=black, colback=gray!10, sharp corners=south, boxrule=0.1pt] \small{\textbf{Takeaway:} ZS is better suited for tasks that are natively query-agnostic and FS is better when dealing with examples such as in QA tasks.}\end{tcolorbox}

\stitle{What Are the Benefits of Integrating Query-Aware Compression Signals?} Integrating query-aware signals (as in KVCompress FS+Q) yields substantial performance gains, particularly in scenarios where computation time is less critical, e.g., test-time settings like Deepseek r1 or OpenAI o1~\citep{guo2025deepseek}. By embedding explicit query cues during compression, the model better prioritizes critical information, as shown by improved accuracy across all tasks and datasets.

\begin{tcolorbox}[colframe=black, colback=gray!10, sharp corners=south, boxrule=0.1pt]\small{\textbf{Takeaway:} Query-aware compression boosts accuracy at the cost of an increase in processing time. Query-agnostic methods like our proposal are the fastest.}\end{tcolorbox}

All results and takeaways are consistent and held true when we applied the same methods and datasets to \textsc{Qwen2.5-7B-Instruct}~\citep{yang2024qwen2}, with its context window extended to 128k using the Yarn technique~\citep{peng2023yarn}, thereby reinforcing the robustness and generalizability of our approach across LLMs.

\begin{figure}[t]
    \centering
    \includegraphics[width=\columnwidth]{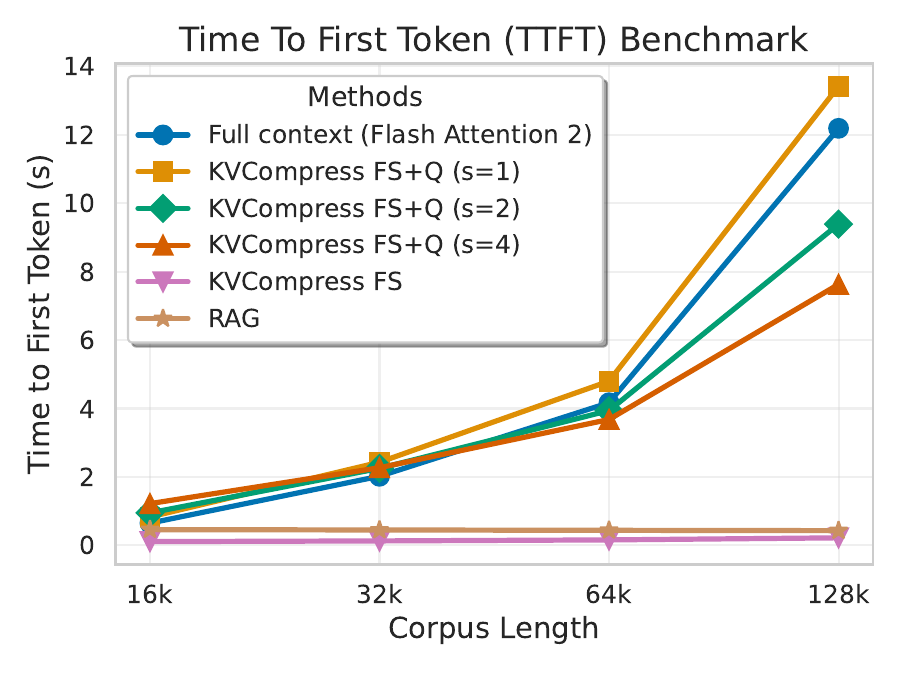}
    \caption{Time to first token with increasing corpus length (context length=8192 and question length=512).}
    \label{fig:ttft}
\end{figure}
\section{Efficiency}

Figure~\ref{fig:ttft} reports inference efficiency for increasing corpus size from 16k to 128k tokens, retrieval context size fixed at 8k tokens, prompt length at 512 tokens, and chunk lengths determined by dividing the corpus length by the number of slices ($s$) as defined in Equation~\eqref{eq:segment_ratio}.

For large corpus lengths, \textsc{KVCompress FS+Q} shows better inference speed compared to processing the entire context with a \textit{Flash Attention 2} implementation~\citep{dao2023flashattention}.
As expected, \textsc{KVCompress FS} %and \textsc{RAG} baselines. \textsc{KVCompress FS} 
exhibits the lowest inference latency across all corpus lengths, being up to 2x faster than \textsc{RAG}. This superior performance arises from the offline precomputation of the KV cache in \textsc{KVCompress FS}, %due to \textsc{KVCompress FS}'s offline KV cache precomputation, 
which enable the model to bypass tokenization and prefill at inference time.

\section{Conclusion and Future work}
%TAG: answering natural language questions over databases~\cite{biswal2024text2sqlenoughunifyingai}. 

In this paper, we presented a task-aware %, query-agnostic
context compression approach that enhances the ability of LLMs to consume large corpora by efficiently populating the KV cache with condensed contextual representations. Our method enables multi-step reasoning while mitigating the limitations of traditional retrieval mechanisms. Empirical results show that our approach outperforms RAG baselines, reduces infrastructure overhead, and improves inference latency.
By distilling raw text into compact representations offline, our method establishes a new paradigm for corpus-level reasoning, addressing critical constraints in existing long-context models. %The compressed cache provides a global view of the corpus, allowing LLMs to synthesize information from extended contexts with near full-context effectiveness but significantly lower computational costs.

%While our work has laid the groundwork for task-aware context compression, several avenues remain unexplored. Firstly, adaptive compression strategies that dynamically adjust based on evolving query characteristics and domain requirements could further enhance the efficiency and effectiveness of our approach. Exploring the use of context compression in real-time scenarios, such as interactive dialogue systems where prompt variability is high, is another direction. Finally, advancing the understanding of compression's impact on semantic integrity would provide critical insights into this transformative approach.

Future directions include integrating head-wise and layer-wise compression strategies, leveraging prior findings that certain heads and layers contribute less to model performance and can be selectively compressed in favor of more informative ones~\citep{feng2024ada, zhang2024simlayerkv}. Additionally, our results highlight a complementary strength between KV compression and RAG: KV compression excels in broad queries, while RAG is more effective for narrow queries. This raises the question of whether a hybrid approach could further enhance retrieval—compressing the corpus offline for global coverage while dynamically fetching top-K chunks online to better address narrow queries. Exploring these strategies could unlock new efficiencies in long-context processing.

\bibliography{tacl2021}
\bibliographystyle{acl_natbib}

\end{document}